%% file: main.tex
\pdfoutput=1
\documentclass{article}
\usepackage{iclr2027_conference,times}
\iclrfinalcopy
\input{math_commands.tex}

\input{tables/numbers}

\usepackage{hyperref}
\usepackage{xcolor}
\definecolor{linkblue}{HTML}{0072B2}
\definecolor{citeorange}{HTML}{B35A1F}
\hypersetup{colorlinks=true,linkcolor=linkblue,urlcolor=linkblue,citecolor=citeorange}
\usepackage{url}
\usepackage{booktabs,multirow}
\usepackage{graphicx}
\usepackage{amsmath,amssymb}
\usepackage[capitalize,noabbrev]{cleveref}
\renewcommand{\floatpagefraction}{0.8}
\makeatletter
\newcommand{\keepwithnext}[1]{\par\vskip\z@\@plus #1\penalty-100\vskip\z@\@plus -#1}
\makeatother

\title{Who Should Score the Cache? \\
Mid-Depth Layers Are the Best Judges \\
of KV-Cache Eviction}

\author{Ekaterina Seregina \\
Colby College \\
\texttt{eseregin@colby.edu}}

\begin{document}
\maketitle

\begin{abstract}
Key-value (KV) cache eviction methods rank tokens by attention scores, and
each layer scores its own cache. To our knowledge, no eviction method asks
which layer should score. We find, across twenty-one models, that the depth of
the scoring layer governs eviction quality. On the eighteen models with
per-layer caches, spanning 0.5B to 109B, long-context retrieval accuracy
collapses at shallow layers, peaks between 0.4 and 0.75 of depth, and degrades
at the last layer. The curve arises because a layer judges tokens well only
once it has used the question to find the answer, which takes several layers.
Modeling eviction as a top-$k$ cut, we compute answer survival probability
from prefill attention and reproduce the measured curves. This approach
selects the scoring layer using calibration prompts with known answers, and
the selected layer is the best layer measured on most models and never
significantly below it.
It improves on the own-layer default on sixteen of the eighteen models, with
significant gains on seven, all at 14B or below, while on larger models the
default is already close to the best mid-depth layer. Scoring every cache from the selected layer
while keeping one set of tokens for all heads also beats per-head SnapKV by 18
to 43 points on five models up to 14B under tight budgets. This gain carries
over to financial filings at 32K context, all at the same memory budget, with
no extra forward pass. The same curve explains a penalty in shared-cache models, where
several layers read one cache and the layer that writes it scores by default.
On Gemma-4-E2B the writer is shallow and loses 28 points to a mid-depth reader
at a 10\% budget. We also find that in open-book question answering
(LongBench) no mid-depth layer beats the own-layer default.
\end{abstract}

\section{Introduction}
The KV cache is the memory cost of long-context inference, and eviction is
the simplest way to shrink it: score every cached token, keep the top $k$,
drop the rest. Existing methods differ in how they score and how they
spend the budget \citep{h2o,snapkv,adakv,pyramidkv}, but all take scores
from the layer whose cache is pruned: layer 7 judges what layer 7 keeps.
The scoring layer is a hyperparameter that, to our knowledge, no eviction
method varies or reports.

\begin{figure}[t]\centering
\includegraphics[width=\linewidth]{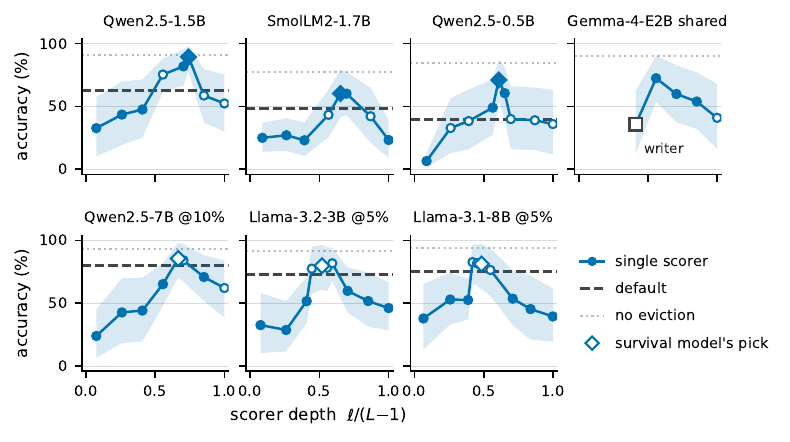}
\caption{\textbf{The depth of the scoring layer governs eviction quality.}
Accuracy when a single layer at depth $\ell/(L{-}1)$ scores every
cache's eviction. All panels: question-window scoring
(\cref{sec:setup}), the eight-task RULER pool, $n{=}200$ (25 per
task). Bands: each layer's task-clustered 95\% CI. Filled markers: the
paired difference from the default excludes zero (task-clustered 95\%
CI, narrower than the bands). Diamonds: the survival model's pick (\cref{sec:model}). Curves
include off-grid layers near the peak.
\emph{Top row:} the four
models whose best mid-depth scorer beats the default, at 15\%; top
right, a shared-cache model, where the default is the
writer ($\square$). \emph{Bottom row:} the three models whose gain
is not significant, each at its binding budget; the curve keeps its
shape.}
\label{fig:curve}
\end{figure}

We find that the chosen layer's depth governs eviction quality at
every scale we test. Accuracy plotted against scorer depth follows the same curve on
every model (\cref{fig:curve}): shallow scorers fail, the last layer is unreliable,
and the best single scorer sits at mid-depth. The mechanism is simple: a
layer judges evictions well once it has used the question to find the answer,
while shallow layers have not yet done so. Building on this idea, we propose a
predictive model with no free parameters that reproduces the measured curves and
selects the scoring layer.

The choice matters most where no own-layer default exists. On models with
cross-layer KV sharing, hereafter \emph{shared-cache} models
\citep{cla,yoco,gemma3n,gemma4}, one cache serves several reader
layers and the scores come by construction from the cache's
\emph{writer}, its shallowest reader. A shallow writer scores from
the region where the curve collapses.

\keepwithnext{4\baselineskip}
\paragraph{Contributions.}
\begin{enumerate}
  \item \textbf{The depth curve.} We measure the depth curve on
  eighteen per-layer models (0.5B--109B, four labs; two repeated at
  32K context; \cref{sec:setup}): scoring quality peaks at a
  middle depth that is stable within a family (0.64--0.75 on
  Qwen2.5 from 14B up; 0.42--0.59 on Mistral and Llama-3). Shallow scoring is
  catastrophic ($-25$ to $-56$ percentage points, pp), and the own-layer default never
  significantly beats the best mid-depth scorer. We find the same curve
  on real documents: on financial-filing extraction, the layer we select on
  RULER gains 8 and 12pp over the default on Llama-3.1-8B and
  Qwen2.5-7B (\cref{sec:pressure}).
  \item \textbf{A better SnapKV.} We change two things in SnapKV, and
  each helps on its own in most settings: scoring from a mid-depth layer, and keeping one
  set of tokens for all heads. Together they beat per-head SnapKV (as
  implemented in the kvpress library) by $+18$ to $+43$pp at budgets of 5 to 15\%
  on five of the seven per-layer models we test (0.5B through 14B)
  and tie it on the two Llama models (\cref{sec:faithful}). On financial filings
at 32K they beat it on both families, by 14 and 16pp. From 32B up we
detect no gain over the default, which is already close to the best
mid-depth layer (\cref{sec:pressure}).
  \item \textbf{A mechanism and a predictive model.} Without the
  question, every layer is nearly equally blind to the answer; with
  it, attention on the answer builds over depth and peaks at mid-depth. We compute the
  probability that the answer survives the scorer's top-$k$ cut, and it
  reproduces the measured curves and the reader ranking on the
  shared-cache model. It also selects the scoring layer: the pick
  is the best layer measured on fourteen of the eighteen per-layer models
  and never significantly below it on the other four (\cref{sec:model}).
  \item \textbf{The shared-cache penalty.} On Gemma-4-E2B
  the shallow writer scores the shared cache by default and
  needs roughly twice a mid-depth reader's budget. A sibling design whose writer sits at mid-depth shows no
  significant penalty, so the writer's depth is the
  likely cause. The penalty replicates across
  budgets, tasks, context lengths, checkpoints, and on Gemma-3n
  (\cref{sec:shared}).
\end{enumerate}

\section{Related Work}
\paragraph{Attention-scored eviction.} These methods score with accumulated attention, recent-window attention, adjusted
softmax, or analytic estimates, and differ in how they spend the budget
\citep{h2o,scissorhands,keyformer,snapkv,tova,adakv,expectedattention,sagekv}.
Within this family the scoring layer never differs: each method scores a cache with the
layer it belongs to. SqueezeAttention, PyramidKV, CAKE, LAVa, RippleKV, and MOSAIC vary
each layer's \emph{budget} while keeping own-layer scores
\citep{squeezeattention,pyramidkv,cake,lava,ripplekv,mosaic}; we compare against
PyramidKV in \cref{sec:faithful}.
\paragraph{Mid-layer token selection.} The closest prior work is GemFilter
\citep{gemfilter}, which selects tokens with an early-middle layer's
attention and then re-runs the model on the shortened prompt. Its layer
ablation was the first to show a middle-layer peak, in that recompression setting;
ASL \citep{asl} picks the selection layer per input there. Multimodal
models prune visual tokens the same way, with a middle layer's attention,
and MAP \citep{map} picks that layer per input.
FastKV \citep{fastkv} forwards only the tokens a mid-depth layer selects to the
layers after it, and LazyLLM \citep{lazyllm} prunes tokens layer by layer
during prefill; both speed up prefill and leave each layer scoring its own cache.
Eviction is a different setting: no second forward pass, one memory
budget for every method, and a default (each layer scoring its
own cache) with no counterpart in recompression. The
finding that this default loses is new.
\Cref{sec:faithful} reports a
head-to-head at the same decode-time memory. HySparse and LongCat Sparse Attention hard-code the
choice at training time: each sparse layer takes a preceding
layer's selections \citep{hysparse,longcat}, which is the writer's role in
\cref{sec:shared}. At decode time, TidalDecode
\citep{tidaldecode} and InfiniGen \citep{infinigen} also move
selection across layers; neither asks which depth selects best.
\paragraph{Cross-layer KV sharing and compression.} CLA, YOCO, LCKV, and
the Gemma-3n and Gemma-4 line made one cache serve many layers
\citep{cla,yoco,lckv,gemma3n,gemma4}. That is the regime this paper
studies. xKV and CommonKV exploit cross-layer redundancy for compression
after the fact \citep{xkv,commonkv}. None of them decides what an
architecturally shared cache should evict.
\paragraph{Per-head differences.} RazorAttention, DuoAttention, FastGen, and HeadKV show
that, within a layer, different heads need different tokens retained
\citep{razorattention,duoattention,fastgen,headkv}. DuoAttention
calibrates its heads on synthetic prompts with known answer spans,
the style our layer selection reuses at layer level (\cref{sec:model}). A shared cache with multi-query
attention removes that freedom: its readers share one physical
cache, which can hold only one set of kept tokens. Per-head selections also
carry a memory cost of their own: stored as masks they keep the full cache,
and heads sharing one buffer keep the union of their picks
\citep{trustmass,pageentrokv}. \Cref{tab:faithful} shows that a
single set scored from the right layer beats per-head selection even
where per-head budgets remain possible.

\section{Setup}\label{sec:setup}
\paragraph{Scoring.} The importance of context token $j$ under scorer
layer $\ell$ is the mean softmax attention that a window of queries
places on token $j$. The depth curves and the shared-cache results
take that window from the question block and average over heads
(the \emph{question-window} convention). The baseline
comparison of \cref{sec:faithful} instead follows the kvpress
implementations, under their own window. The
same scorer at the same budget shifts by several points between the
two windows, so we never compare numbers across the two windows.
\paragraph{Eviction.} A \emph{budget} $b$ keeps $k=\lfloor b\,T_c
\rfloor$ of a prompt's $T_c$ context tokens in every layer's KV cache and
masks the rest. Some tokens are always kept, and they count against $k$. Under
the question window these are the sink (first 4) and the recency window
(last 32) \citep{streamingllm}; under kvpress's window, the scoring window
itself. All methods
compete at the same $k$. Each model is reported at its \emph{binding
budget}, the smallest candidate budget at which the own-layer default
loses at least 5pp to the uncompressed model while no task collapses
(default recall under 5\%): eviction costs accuracy but leaves signal to
compare scorers. At these budgets the default has lost 7 to 41pp (\cref{app:ledger-binding}). The three weakest primary models (15\%)
and the Gemma-4 shared cache (15\% in \cref{fig:curve}, 10\% in \cref{tab:shared}) are shown at uniform budgets; their binding budgets are 15 to 25\%, and the gains hold there (\cref{app:ledger-binding}).
We compress after prefill, so the savings apply to the cache that
persists through decode. We mask evicted context
positions at decode ($-\infty$ additive mask, per-layer and
per-head where the baseline requires). Masking preserves RoPE
positions and is mathematically equivalent to deleting the rows
(\cref{app:physical}).
\paragraph{Models, tasks, metrics.} We study twenty-one models: eighteen
with per-layer caches and three with shared caches (Gemma-4-E2B,
Gemma-4-E4B, Gemma-3n-E2B). The primary set is six instruction-tuned
per-layer models (Qwen2.5-\{0.5B,1.5B,7B\}, SmolLM2-1.7B, Llama-3.2-3B,
Llama-3.1-8B) and Gemma-4-E2B (base checkpoint; the instruct one replicates,
\cref{app:ledger-gemma}), where the writer plays the ``own layer'';
twelve more per-layer models extend the range of scale and release date
under the same setup (\cref{tab:scale}).
The benchmarks are RULER \citep{ruler} at 4K, 8K, 16K, and 32K,
LongBench v1 \citep{longbench}, and
DocFinQA \citep{docfinqa}, numeric questions over 10-K filings
(\cref{app:ledger-docfinqa}). Every primary RULER result pools the
same eight RULER-4096 tasks (five needle-retrieval variants, variable
tracking, two QA) and, for each model, counts only the informative ones, those the
uncompressed model gets at least 30\% right (all eight, on every per-layer model). Adding the other five RULER tasks shrinks
every gain and does not change any conclusion (\cref{app:ledger-tasks}). Decoding is greedy.
RULER is scored as recall: the fraction of a prompt's gold answers that
appear in the output. The LongBench and DocFinQA metrics are in
\cref{app:ledger-longbench,app:ledger-docfinqa}.

\paragraph{Statistics.} Results use 25 prompts per task, $n{=}200$ for
the eight-task pool. Prompts are nested in tasks, so all primary
confidence intervals are two-stage cluster bootstraps (resample tasks,
then prompts). Bold numbers and filled markers mean the
paired difference's 95\% interval excludes zero
(clustered-significant, for short). Percentile bootstraps
can undercover at $K{=}8$ tasks, so the six primary comparisons
are also checked with an exact sign-flip test
(\cref{app:ledger-signflip}). No multiple-comparison correction is applied.
\Cref{app:ledger} records how each budget was set and how each pick performed.

\section{The Depth of the Scoring Layer}
\Cref{fig:curve} is the paper's central result. We let one layer at a
time score every cache and plot accuracy against that layer's depth.
On each per-layer model we test seven grid layers, at depths 0.08,
0.25, 0.40, 0.55, 0.70, 0.85, and 1.0, plus extra layers near the peak,
including the survival model's pick when it is not a grid layer. The
curve has the same shape on every model: a cliff on the shallow side,
a peak at 0.4--0.75 of depth, and a decline toward the last layer. The
default loses to the best mid-depth layer on all seven models,
significantly on the four in the top row.
The diamonds mark the \emph{survival model's pick} (\cref{sec:model}),
chosen from prefill attention and known answer locations on
calibration prompts. The picked layer beats the default by $\svSmolOwn$pp
(SmolLM2), $\svOneFiveOwn$pp (1.5B), and $\svHalfOwn$pp (0.5B) on the
top row's three per-layer models, and on the
shared-cache panel the best reader recovers $\gPanelGain$pp over the
writer, all
clustered-significant. The three models whose gain is not significant
(3B, 7B, 8B) are considered in \cref{sec:pressure}.

\subsection{Against the existing methods}\label{sec:faithful}
The depth curves use our question-window scoring. \Cref{tab:faithful}
re-tests the gains inside SnapKV as kvpress runs it: the last 64 prefilled tokens are the query
window (\cref{app:physical}), scores are smoothed over 5 tokens, and each
head keeps its own token set. All methods share one memory budget. The
last three methods change SnapKV's two defaults, one at a time and then
together: one kept set for all heads, scoring from the survival model's
pick, and both.

On its own, the mid layer helps in every column and the single set in
every column but the 14B (\cref{tab:ledgerparts}). The combination scores
above either alone in every column. It beats per-head SnapKV by $+18$ to $+43$pp,
clustered-significant in every column (exact tests in
\cref{app:ledger-signflip}), and recovers 83 to 99\% of the uncompressed
accuracy where SnapKV keeps 37 to 63\%. Beyond these columns the combination wins on
Qwen2.5-0.5B and ties on the two Llama models, whose depth gains are not
significant either (\cref{tab:ledgertabone}). On the 1.5B the lead holds
from 5\% to 50\%: it peaks at 10\%, then shrinks as SnapKV recovers, and stays significant
throughout (\cref{app:ledger-budgets}).

AdaKV, which redistributes budget across heads using SnapKV's scores,
stays within 4.8pp of SnapKV in these columns and differs significantly in one
(\cref{tab:ledgerparts}). PyramidKV shrinks per-layer budgets with depth,
following its authors' schedule. At the same total memory it scores
\emph{below} SnapKV in every column of \cref{tab:faithful} ($-8.2$ to
$-29.4$pp): PyramidKV's schedule costs more here than ignoring depth.
\begin{table}[t]
\caption{\textbf{Changing SnapKV's scoring layer and dropping its per-head
selection each help, and together help most.} Accuracy (\%) on the
eight-task pool of \cref{sec:setup}, $n{=}200$ per column; 1.5B, 7B, and
14B are Qwen2.5. Every row but H2O scores with the kvpress window, so numbers are
not comparable to \cref{fig:curve}. Single set: one kept set for all
heads. Mid-layer: scored from the survival model's pick at the column's
budget (L20, L20, L15, L18, L30; \cref{sec:model}); in its per-head row, KV
head $h$ of the mid layer selects for KV head $h$ of every layer. Columns
are separate runs, each with its own ceiling. Bold accuracy: best method
in the column. $\Delta$ row: task-clustered 95\% CI; bold $=$ significant.}
\label{tab:faithful}
\centering\footnotesize
{\setlength{\tabcolsep}{1.4pt}\input{tables/tab1_faithful}}
\end{table}

\paragraph{Recompression head-to-head.} GemFilter \citep{gemfilter},
the nearest competing method, selects tokens with one layer's
attention and re-runs the model on them alone; its layer is fixed at
a set fraction of depth (L11 here). With the same decode-time cache size
(Qwen2.5-1.5B, 15\%), it reaches $\gfGem\%$ against our $\gfOurs\%$
(L20 under the question window, as in \cref{fig:curve}; ceiling $\gfCeil\%$; clustered-significant). The layer explains the
gap: GemFilter moved to L20 reaches $\gfGemPeak\%$, within noise of
the uncompressed model and of our eviction, which needs no
second prefill. GemFilter targets far longer prompts; the comparison
holds in our setting only.

\section{The Model: Eviction as Top-$k$ Selection}\label{sec:model}
\begin{figure}[t]\centering
\includegraphics[width=\linewidth]{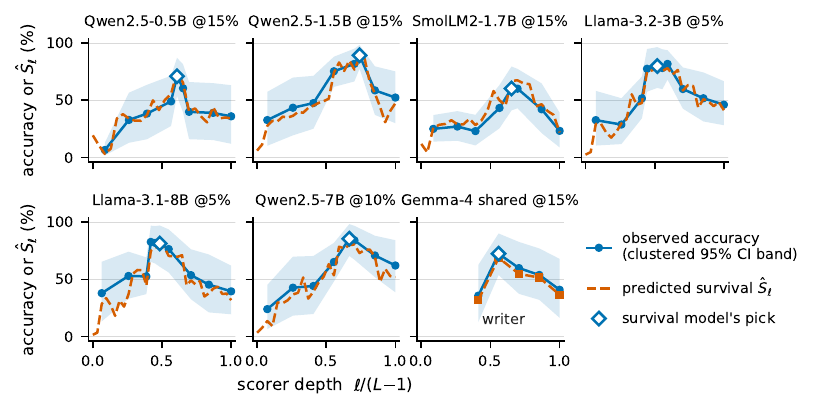}
\caption{\textbf{The survival model predicts the depth
curve.} Blue: observed accuracy at the measured layers, on and off the
seven-layer grid, with task-clustered 95\% CIs. Orange dashed: predicted
survival $\hat S_\ell$ (\cref{eq:surv}) from the uncompressed model's
prefill attention, computed at every layer. Diamonds: the survival
model's pick, at its observed accuracy.
\emph{Last panel:} the same construction on the
Gemma-4 shared cache, one point per reader.}
\label{fig:model}
\end{figure}

Eviction is a ranking problem: when one layer $\ell$ scores every
cache, the kept set is the top $k$ of its scores, so $\ell$ is a good scorer when it ranks the tokens the answer needs above $k$-th place. For
each prompt we reconstruct the set $K_\ell$ of tokens layer $\ell$
would keep, including the always-kept sink and recency tokens. A gold
answer \emph{survives} if some occurrence of it lies
entirely inside $K_\ell$, matching the metric's all-or-nothing credit.
The prediction is survival averaged over a prompt's gold answers, then over prompts,
\begin{equation}
\hat S_\ell \;=\; \frac{1}{n}\sum_{\text{prompts}}\;
\frac{1}{|G|}\sum_{g\in G}
\mathbf{1}\!\left\{\exists\, o \in O_g:\; o \subseteq K_\ell\right\},
\label{eq:surv}
\end{equation}
where $G$ is the prompt's gold set and $O_g$ the token-index sets of gold
$g$'s occurrences. Everything on the right side comes from the prefill
attention of the uncompressed model and the locations of the gold
spans. It needs no generation and has no adjustable constants; we
call it the \emph{survival model}. It is computed on 25 calibration
prompts per task, drawn from dataset rows the evaluation never uses.

\Cref{fig:model} overlays $\hat S_\ell$ on the depth curves. The
survival model's pick is the layer with the highest $\hat S_\ell$. When the
top two layers differ by less than one standard error of their paired
per-prompt difference, $\hat S_\ell$ cannot separate them, and the pick is the
shallower of the two: $\hat S_\ell$ counts only the answer span, and a deeper
scorer can drop other context the model needs. The pick (diamonds
in \cref{fig:model}) is the best layer measured on fourteen of the
eighteen per-layer models and never significantly below it on the other
four (\cref{tab:ledgerpick}). On the six per-layer models in \cref{fig:model},
$\hat S_\ell$ tracks the measured curves with a Spearman correlation of
$\spLo$--$\spHi$ and a mean absolute error of $\maeLo$--$\maeHi$pp. We also compute $\hat S_\ell$ from each reader's attention
over the Gemma-4 shared cache; it reproduces all five readers'
accuracies within $\gModMax$pp ($r{=}\gModR$; \cref{fig:model}, last
panel), so the per-layer curves and the shared-cache penalty are one
phenomenon.

\subsection{The mechanism}\label{sec:mechanism}
Our explanation for the middle peak is that the network needs several
layers to find the answer with the question. Early layers have not
finished that work, so their attention cannot yet pick out what the question needs.
The last layer has moved on to producing output. The layers in between
use the question to retrieve, so their attention ranks tokens well. The cliffs at both
ends come from the all-or-nothing rule: an answer survives only if every
one of its tokens is kept, so moderate ranking errors add up to lost
answers.

\begin{figure}[t]\centering
\includegraphics[width=\linewidth]{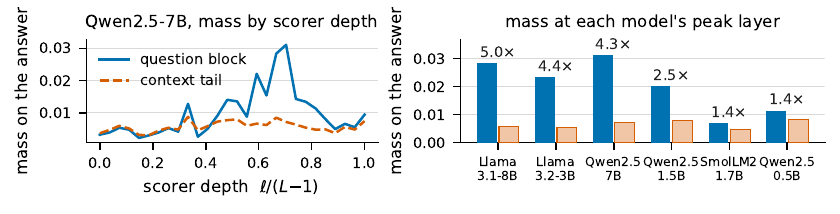}
\caption{\textbf{Taking the question away flattens the depth signal.}
Attention mass on the answer's tokens, scored by the question block
(blue) or the context's last 32 tokens (orange); means over the 199
prompts with located answer spans, no eviction. \emph{Left:} Qwen2.5-7B by scorer depth.
\emph{Right:} mass at each model's peak layer, with and without the
question; the number above each pair is their ratio.}
\label{fig:noq}
\end{figure}

The direct test is to take the question away. Every depth-curve score
in this paper comes from attention paid by the question's tokens
(\cref{sec:setup}). For the test we compute the same attention from
the context's own final tokens instead, as if no question had been
asked, and compare how much attention mass each layer places on the
answer's tokens. With the question, Gemma-4's reader layers place
$\gMassRatio\times$ as much mass on the answer as the writer does.
Without it, the readers and the writer are nearly equally blind to the
answer, and the gap between them shrinks by a factor of $\gGapShrink$.

\Cref{fig:noq}
shows the same collapse on the per-layer models. On the two weakest models, Qwen2.5-0.5B and
SmolLM2, the drop is small: there the needle draws much of its
attention simply by standing out from the surrounding text, with or
without the question. With the question, the mass on the answer peaks
at mid-depth, within one grid step of the accuracy peak, on every model
we test.

\paragraph{Heads versus depth.} Retrieval heads concentrate at mid-depth
\citep{retrievalhead}, so the depth curve could simply reflect where
they sit. As a control, we score every cache with the pooled attention
of the sixteen heads with the most attention mass on the answer, chosen
on two calibration prompts per task. The head scorer beats the default
on all three models tested. It ties the best layer on Qwen2.5-1.5B and
Llama-3.1-8B and beats a mid layer on Mistral-Small-24B, where good and
poor layers interleave (\cref{app:ledger-rh}). The good heads live at
mid-depth on every model. A single layer collects them without a head
search, and choosing heads within that region would refine the pick.

\subsection{Where the advantage holds and where it ends}\label{sec:pressure}
The three models of \cref{fig:curve}'s bottom row show no significant
pooled gain. On the 7B and the 8B the default already ties the
mid-depth plateau. On the 3B, gains on the multi-needle tasks offset
losses on variable tracking and QA.

\paragraph{Scale.} The mid-depth gain persists to 14B and is undetectable from
32B (\cref{tab:scale}). Significant gains of $+12$ to $+31$pp below 2B continue
at 14B ($\svFourteenOwn$pp at the pick, L30). From 32B up the
survival model's picks (32B, 70B, 72B, 109B) tie the default too
(\cref{tab:ledgerpick}), so the grid did not miss a better layer.
The cost of a wrong scorer does not shrink: the worst layer loses 25
to 56pp at every scale. This matters when the architecture
fixes the scorer (\cref{sec:shared}) or a deployment adopts one scorer
without checking it.

\begin{table}[t]
\caption{\textbf{Scale, release date, and context length: the win shrinks,
the downside stays.}
Best and worst single scorer among the layers tested, against the
per-layer default (own), each model at its binding budget; $n{=}200$, task-clustered 95\% CIs, bold
$=$ significant. The survival model's picks are in \cref{tab:ledgerpick}. The 70B and 72B run
in 4-bit NF4. Last two rows: 32K contexts.}
\label{tab:scale}
\centering\footnotesize\setlength{\tabcolsep}{2.5pt}
\input{tables/tab3_scale}
\end{table}

\paragraph{Newer families, hybrid attention, and MoE.} The curve and
the shrinking margin replicate on the 2025--2026 releases
(\cref{tab:scale}). On Qwen3 all three sizes peak at a
middle depth, and the margin falls with capability: significant at 0.6B, not significant at
1.7B, gone at 14B. The curve also holds on Qwen3.5-9B, where gated linear
attention leaves only 8 of 32 layers with a KV cache, and on the
mixture-of-experts Llama-4-Scout (109B), whose layers without positional
encoding score like their neighbors (\cref{app:ledger-picks}).

\paragraph{Task family.} The curve carries to real documents when
exact values must survive the cut. On DocFinQA \citep{docfinqa}, the
layer selected on RULER at 32K (the \emph{RULER pick}) beats the
default significantly on both Llama-3.1-8B and Qwen2.5-7B, by 8 and
12pp, and each pick lies at its family's usual depth
(\cref{app:ledger-docfinqa}). Against per-head
SnapKV the combination of \cref{tab:faithful} wins there too, by $\lcDfqEight$ and
$\lcDfqSeven$pp (\cref{tab:ledgerlong}).
Open-book question answering keeps the curve's
ends but loses its gain: on LongBench the shallow and last layers
still lose, by 5 to 6 F1, and no mid layer beats the default
(\cref{app:ledger-longbench}). The survival model, which was built for
exact-match answers, does not predict that outcome. The RULER pool splits the same way at 4K and at 32K:
the margin over SnapKV comes from the needle tasks and variable tracking; the
two QA tasks add little on the small models and lose up to 12pp on the larger ones (\cref{tab:ledgerpertask}).

\paragraph{Context length.} At 32K the curve keeps its shape and its
family depth: Llama-3.1-8B and Qwen2.5-7B peak at their picks, L14 and L19. The
shallow third and the last layer sit 20 to 48pp below the default, and
the best layer ties it (\cref{tab:scale}). \Cref{tab:faithful}'s comparison repeats its
4K outcome: the combination beats SnapKV by $\lcRulerSeven$pp on Qwen2.5-7B and ties it on
Llama-3.1-8B (\cref{tab:ledgerlong}). Eviction binds only at
0.5\% and 1\%, which keep 160 to 330 tokens, about what the 4K budgets
keep (\cref{tab:ledgerbind}): pressure depends on the number of kept
tokens rather than their fraction of the context.

\section{Shared Caches: When the Architecture Picks the Scorer}\label{sec:shared}
\begin{table}[t]
\caption{\textbf{On Gemma-4-E2B the default scores from the shallow
end of the curve.} Layer-14 cache evicted to
a 10\% budget; $n{=}200$.
The writer needs a 15--25\% budget to match reader L19 at 10\% (\cref{tab:ledgerbind}). $\Delta$: task-clustered 95\% CIs; bold $=$
significant.}
\label{tab:shared}
\centering\footnotesize
\input{tables/tab2_shared}
\end{table}
Gemma-4-E2B computes one KV cache at layer 14 and reads it from layers
14, 19, 24, 29, and 34. The five layers share one physical cache, so it
can hold only one set of kept tokens, and the architecture decides who
scores it: by default the writer, the layer that computes the cache,
ranks the tokens with its own attention. On E2B the writer is layer 14
of 35 (depth 0.41), the shallowest of the five readers, and it sits on
the shallow side of the curve (\cref{fig:curve}); the deeper readers
place $\gMassRatio\times$ as much question-driven mass on the answer as
it does (\cref{sec:mechanism}). We evict the shared cache only and leave
the private caches intact (\cref{app:ledger-gemma}).

\Cref{tab:shared} shows what the default costs. At a 10\% budget the
writer scores 23\% where reader L19 scores 51\%, and every reader beats
the writer with clustered significance. Rank consensus over the
readers, which keeps the tokens with the best mean rank and removes the
need to pick a layer, recovers most of L19's gain.
The penalty replicates across budgets, context lengths, prompt draws, and
checkpoints (\cref{app:ledger-gemma}). It is smaller on Gemma-3n, whose
writer sits at 0.66 and whose best reader sits deeper still, at 0.83
($+12.5$pp).

\paragraph{The penalty follows the writer's depth.} Gemma-4-E4B, a
sibling design, places its writer at depth 0.56. On E4B the penalty
largely disappears: the $+22.2$pp consensus-over-writer gain of
\cref{tab:shared} shrinks to $\eFourGain$pp $\eFourGainCI$ at the same
10\% budget, and the task-paired difference between the two designs is
$\xDesign$pp $\xDesignCI$. Beyond retrieval the evidence weakens. On
LongBench (seven datasets) consensus beats the writer on E2B-it by
$\lbEtwo$ F1 $\lbEtwoCI$, but the two designs do not differ
significantly there. On DocFinQA, E4B-it's mid-depth writer beats every
deeper reader by 14 to 22pp (\cref{app:ledger-docfinqa}): when the
writer already sits at mid-depth, the architecture has picked well.

\section{Practical Guidance, Cost, and Limitations}\label{sec:practical}
\label{sec:limitations}On per-layer models, let the survival model
choose the scoring layer (\cref{sec:model}) and let that layer score
every cache with a single set of kept tokens. Do not average across all layers:
rank consensus over all layers loses 21--37pp to the selected layer on two models
(\cref{app:ledger-consensus}).
On shared-cache models with a shallow writer, score with a mid-depth
reader or with rank consensus over the cache's readers. Scoring reads one
layer's attention at prefill and adds under 1\% to its time
(\cref{app:physical}). Physical
eviction at a 1\% budget shrinks Llama-3.1-8B's 32K
decode-resident cache from 4.0GB to 45MB and speeds decode $1.4\times$
(\cref{app:physical}).

The evidence has limits. The pick needs calibration prompts with known
answers. The shared-cache results come only from Gemma models, the scale
trend on newer releases only from Qwen3, and the 70B and
72B runs from 4-bit NF4 checkpoints. Three problems remain open. First,
the best scorer differs by task family (\cref{sec:pressure}), so a selector that
detects the task family at prefill could beat any single layer. Second, the
writer-depth explanation rests on one sibling pair (\cref{sec:shared}).
Training the same model with a deeper writer would test it directly. Third, every scorer here sees the question, and SnapKV's edge
disappears without it \citep{queryvis}, so prefix caches need a
question-free signal \citep{kvzip}.\label{end:limits}

\subsubsection*{Acknowledgments}
Computations were run on the Colby College High-Performance Computing cluster.

\subsubsection*{Reproducibility statement}\label{repro:start}
RULER prompts at 4K--16K come from the \texttt{simonjegou/ruler} Hugging
Face release and at 32K from lighteval's official-generator releases, one
per tokenizer (e.g.\ \texttt{lighteval/RULER-32768-Lamma3-Instruct}).
Models run in fp16, except Llama-4-Scout and Qwen3.5-9B in bf16 and the
70B and 72B in 4-bit NF4. Prompt draws are deterministic. Greedy decoding is
bit-exact only within one GPU and kernel configuration: rerunning the
primary sweeps on a different GPU flipped a few near-tied prompts per
configuration and moved pooled accuracies by at most a point. Every
number is computed by a script and stored in a JSON
file; figures, tables, and the statistics quoted as macros read these
files, and the prose numbers come from those tables and files. One
statistics implementation is shared by every figure and table. Baseline semantics were ported from the
kvpress reference implementations (\url{https://github.com/NVIDIA/kvpress},
version 0.5.4; PyramidKV's schedule from its authors' code), and
\cref{app:physical} verifies that masked eviction matches physical
deletion. On the same prompts, our SnapKV scoring and kvpress agree on
over 99.8\% of kept tokens (Llama-3.1-8B).
\label{end:maintext}

\bibliography{refs}
\bibliographystyle{iclr2027_conference}

\appendix
\setcounter{table}{0}\renewcommand{\thetable}{A\arabic{table}}%
\renewcommand{\theHtable}{A\arabic{table}}
\setcounter{figure}{0}\renewcommand{\thefigure}{A\arabic{figure}}%
\renewcommand{\theHfigure}{A\arabic{figure}}
\renewcommand{\topfraction}{0.95}\renewcommand{\bottomfraction}{0.7}
\renewcommand{\textfraction}{0.05}\renewcommand{\floatpagefraction}{0.9}
\setcounter{topnumber}{3}\setcounter{bottomnumber}{2}\setcounter{totalnumber}{4}
\section{Supporting comparisons}\label{app:ledger}
This appendix records the two choices made for each model, its budget
(\cref{app:ledger-binding}) and its scoring layer
(\cref{app:ledger-picks}), and then the comparisons the main text cites
(\cref{app:ledger-tasks} to \cref{app:ledger-signflip}).

\subsection{Binding budgets}\label{app:ledger-binding}
\Cref{tab:ledgerbind} shows the scan behind every rule-chosen budget in
the paper. The scan uses the first ten evaluation prompts per task and
depends only on how much the default loses. Budgets loosen as models
weaken. Four primary models appear in the main text at a uniform budget
in place of their binding budgets (\cref{sec:setup}); for Qwen2.5-1.5B
the two coincide at 15\%. At the binding budgets the gains hold. The pick beats
the default by $\rbHalfOwn$pp on Qwen2.5-0.5B (25\%) and $\rbSmolOwn$pp on
SmolLM2 (20\%), and the combination of \cref{tab:faithful} beats per-head
SnapKV by $\rbHalfSnap$ and $\rbSmolSnap$pp there, all
clustered-significant; the Gemma-4-E2B gap at 25\% is in
\cref{tab:ledgergemma}.

\begin{table}[!htbp]
\caption{\textbf{How each model's budget was set.} Accuracy lost by
the own-layer default at each candidate budget (uncompressed minus
default, pp, informative tasks, ten prompts per task). Bold: the chosen
budget, the smallest that loses at least 5pp with no task collapsed
($^{\mathrm{f}}$: default recall under 5\% on a task the uncompressed
model gets above 30\%). Dashes: budget not scanned. The candidates are
$\{5,10,15\}\%$, extended when nothing bound: up to 25\% for the weakest
models, down to 2\% for Llama-4-Scout and to 0.25\% at 32K. Blocks: the
primary models (on Gemma-4-E2B the default is the writer), the twelve
models of \cref{tab:scale}, the 32K runs.}
\label{tab:ledgerbind}
\centering\footnotesize\setlength{\tabcolsep}{3pt}
\input{tables/ledger_binding}
\end{table}

\subsection{The survival model's picks}\label{app:ledger-picks}
\Cref{tab:ledgerpick} tests the survival model's pick on every
per-layer model and on two of them at 32K. Against the default the pick
wins significantly on seven models and ties the rest, the largest
models included; the margin is large where the peak is sharp and a tie
where the top is flat. The pick is the best layer measured on fourteen
of the eighteen models and within noise of the best on the other four;
it beats the best other layer significantly on two. Bootstrapping the calibration
prompts leaves the pick unchanged in 66 to 100\% of resamples on the
seven models with a significant gain. Llama-4-Scout drops positional
encoding in every fourth layer; the four such layers that we ran
score like their neighbors (54.7 to 66.6\% against 60.5 to 70.8\%), so
the layer type does not disturb the depth curve.

\begin{table}[!htbp]
\caption{\textbf{The survival model's picks.} Pick: the layer the
survival model selects (\cref{sec:model}) on 25 calibration prompts per
task, disjoint from the evaluation prompts. Best other: the
highest-scoring other layer measured on the model. \emph{Bottom block:} the
same rule at 32K context on two models. $n{=}200$; task-clustered 95\%
CIs; bold: CI excludes zero.}
\label{tab:ledgerpick}
\centering\footnotesize
\resizebox{\linewidth}{!}{\input{tables/ledger_picks}}
\end{table}

\subsection{The task pool}\label{app:ledger-tasks}
\Cref{tab:ledgertasks} checks the five RULER tasks left out of the
eight-task pool (\cref{sec:setup}). Three score low under both methods
and so carry little signal: niah\_single\_3, whose needle is a UUID of
about thirty tokens that the all-or-nothing metric never credits once
one is lost, cwe, and niah\_multikey\_3. On the other two, fwe and
niah\_multikey\_2, ours matches or beats SnapKV. Over all thirteen tasks
every gain shrinks by 3--6pp and stays significant.

\begin{table}[!htbp]
\caption{\textbf{The five excluded RULER tasks.} Accuracy (\%) of the
uncompressed model (none) and of the mid-layer single set and per-head
SnapKV (ours/SnapKV), ten prompts per task. Bottom rows: ours minus
SnapKV, pooled over prompts, on the five excluded tasks (ten prompts
each), the eight-task pool of \cref{tab:faithful} (25 each), and all
thirteen tasks; informative tasks only, as throughout. Task-clustered
95\% CIs; bold: CI excludes zero.}
\label{tab:ledgertasks}
\centering\footnotesize\setlength{\tabcolsep}{2.5pt}
\input{tables/ledger_taskpool}
\end{table}

\subsection{The method comparison beyond Table 1}\label{app:ledger-budgets}
\Cref{tab:ledgertabone} extends the \cref{tab:faithful} comparison to
Qwen2.5-1.5B at five more budgets, to the three remaining primary
models, and to 32K. The combination's lead over per-head SnapKV holds
at every 1.5B budget, shrinking above 10\% as SnapKV recovers, and on the 0.5B;
the two Llama models, whose depth gains are not significant, tie. At
32K it beats SnapKV on both models on DocFinQA and on Qwen2.5-7B on
RULER, and ties on Llama-3.1-8B, as at 4K. \Cref{tab:ledgerparts} splits
the margin into its two changes (top) and by task (bottom).

\begin{table}[!htbp]
\caption{\textbf{The method comparison beyond Table 1.} Rows and
conventions as in \cref{tab:faithful}; the scoring layer is the survival
model's pick at the column's budget. \emph{Top:} Qwen2.5-1.5B at two more
budgets, and the three remaining primary models at the budgets of
\cref{fig:curve}. \emph{Middle:} Qwen2.5-1.5B at looser budgets (picks L16, L16, L20). \emph{Bottom:} at 32K, on the DocFinQA prompts and budgets of
\cref{tab:ledgerdfq} and on the RULER pool at the budgets of
\cref{tab:scale}; H2O was not run at 32K, where its full 32K$\times$32K
attention did not fit. DocFinQA intervals are over
prompts (one task), and its numbers use the kvpress window, so they
differ from \cref{tab:ledgerdfq}'s.}
\label{tab:ledgertabone}\label{tab:ledgerlong}
\centering\footnotesize\setlength{\tabcolsep}{2.5pt}
\input{tables/ledger_tab1_more}\\[6pt]
\input{tables/ledger_tab1_loose}\\[6pt]
\input{tables/ledger_tab1_32k}
\end{table}

\begin{table}[!htbp]
\caption{\textbf{Both changes contribute, and the margin comes from the
needle and variable-tracking tasks.} \emph{Top:} each change on its own,
and AdaKV, against per-head SnapKV in the columns of \cref{tab:faithful};
task-clustered 95\% CIs; bold: CI excludes zero. The 14B rows come from
two runs on the same prompts, each paired with the SnapKV of its own
run. \emph{Bottom:} accuracy of the combination minus per-head SnapKV on
each task (pp, 25 prompts per task and column); each column averages to
its $\Delta$ row in \cref{tab:faithful} or \cref{tab:ledgertabone}.}
\label{tab:ledgerparts}\label{tab:ledgerpertask}
\centering\footnotesize\setlength{\tabcolsep}{2pt}
\input{tables/ledger_tab1_parts}\\[6pt]
\input{tables/ledger_pertask}
\end{table}

\subsection{Consensus over layers}\label{app:ledger-consensus}
Averaging across layers loses to the pick, which backs the advice of
\cref{sec:practical}. The pick beats rank consensus over all layers by
$\consOneAll$pp on Qwen2.5-1.5B and $\consSmolAll$pp on SmolLM2-1.7B, and
consensus over the middle third by $\consOneMid$ and $\consSmolMid$pp, all
clustered-significant (eight-task pool at 15\%, $n{=}200$; the pick, L20
and L15, measured in a separate run on the same prompts). Averaging
dilutes the one sharply good layer with its neighbors.

\subsection{LongBench}\label{app:ledger-longbench}
\Cref{tab:ledgerlb} gives the LongBench result of \cref{sec:pressure}
on two models, pooled over datasets. No scorer beats the default: the
mid layers tie it and the shallow and last layers lose 5 to 6 F1 on
both models. The same Llama-3.1-8B layers gain 4 to 12pp on the filings
(\cref{app:ledger-docfinqa}). LongBench's two code-completion datasets,
LCC and RepoBench-P, were left out because eviction barely touches
them: on Qwen2.5-1.5B the default loses at most 0.5 points of edit
similarity, LongBench's metric for code, at any budget from 2\% to
10\%, so no budget binds.

\begin{table}[!htbp]
\caption{\textbf{On LongBench no mid layer beats the default, and the
shallow and last layers still lose.} Nine question-aware datasets
(HotpotQA, 2WikiMQA, MuSiQue, Qasper, MultiFieldQA-en, TriviaQA,
NarrativeQA, QMSum, PassageRetrieval-en), official templates and metrics
(normalized F1; ROUGE-L for QMSum, the retrieval score for
PassageRetrieval-en), 25 prompts each, pooled over the datasets whose
no-eviction score is at least 15. Qwen2.5-1.5B at 5\%, its binding
budget on this benchmark (L19, the grid mid layer, and L20, the pick);
Llama-3.1-8B at 2\% (L13, L14, L17, the plateau of its RULER curve).
Task-clustered 95\% CIs against the default; bold: CI excludes zero.}
\label{tab:ledgerlb}
\centering\footnotesize
\input{tables/ledger_longbench}
\end{table}

\subsection{Shared-cache replications and the private caches}\label{app:ledger-gemma}
\Cref{tab:ledgergemma} repeats the comparison of \cref{sec:shared} at
other budgets and context lengths, on a disjoint prompt draw, on all
thirteen RULER tasks for both checkpoints, and on Gemma-3n. Every gap
is significant. On LongBench (the seven F1-scored datasets of
\cref{tab:ledgerlb}, 12 prompts each, 10\% budget) consensus beats the
writer by $\lbEtwo$ F1 on E2B-it and by a non-significant $\lbEfour$ on
E4B-it.

Gemma-4-E2B has two other caches that grow with the prompt, written by
layers 4 and 9 and read only by those layers; the main experiments
leave them intact. In a separate run, evicting them to 10\% costs 21pp
when each layer ranks its own tokens and 9pp under reader L19's
ranking, a difference of $+12.3$pp (95\% CI $[+0.2,+25.6]$; sign-flip
$p{=}0.16$). Even as a cache's only reader, a shallow layer scores it
poorly.

\begin{table}[!htbp]
\caption{\textbf{The shared-cache penalty replicates across budgets,
context lengths, prompt draws, and checkpoints, and on Gemma-3n.}
\emph{Top:} reader L19 minus the writer on the shared cache across
budgets and context lengths, in separate runs (Figure 1's panel gives $+36.7$ at 15\%); last row, Gemma-3n's reader L24 minus its
writer L19 (13 tasks, $n{=}325$). \emph{Bottom:} the 10\% comparison on a
prompt draw disjoint from every other Gemma table (five tasks, ten
prompts each), and on all thirteen RULER-4096 tasks (five prompts each,
informative tasks pooled) for the base and instruct checkpoints; cons.:
rank consensus over the five readers. Task-clustered 95\% CIs; bold: CI
excludes zero.}
\label{tab:ledgergemma}
\centering\footnotesize\setlength{\tabcolsep}{3pt}
\input{tables/tab2_shared_reps}\\[6pt]
\input{tables/ledger_gemma}
\end{table}

\subsection{Heads as scorers}\label{app:ledger-rh}
\Cref{tab:ledgerrh} backs the heads-versus-depth test of
\cref{sec:mechanism}. The sixteen (layer, head) pairs with the most
attention mass on the answer, chosen on two calibration prompts per
task, are pooled into one ranking that scores every cache. The three
models are the cases where head selection had the best chance to beat
depth: a sharp peak (Qwen2.5-1.5B), a flat plateau with no significant
depth gain (Llama-3.1-8B), and interleaved good and poor layers
(Mistral-Small-24B). On Gemma-4 the writer's gap in attention mass (\cref{sec:mechanism}) is not an
artifact of averaging over its heads: each reader's best single head
carries $\gBestHeadLo$ to $\gBestHeadHi\times$ as much mass on the answer
as the writer's best head.

\begin{table}[!htbp]
\caption{\textbf{The top-16 retrieval heads beat the default on every
model and tie or beat the mid layer they are compared with.} Eight-task
pool, $n{=}200$, each model at its binding budget; layer: the mid layer
the heads are compared with (on Qwen2.5-1.5B, L19, the grid mid layer,
and L20, the pick). Task-clustered 95\% CIs; bold: CI excludes zero.}
\label{tab:ledgerrh}
\centering\footnotesize
\input{tables/ledger_rh}
\end{table}

\subsection{Financial-filing extraction}\label{app:ledger-docfinqa}
DocFinQA \citep{docfinqa} asks numeric questions about full 10-K
filings. Each question comes with the arithmetic steps that produce its
answer; the gold set is the figures those steps take from the filing,
as they appear there. The model sees a 31.5K-token slice that contains
them at a random offset and is asked to state the figures it uses; we
score recall of those figures. The uncompressed model scores under
50\% on this task, so the binding budget's 5\% collapse threshold is too low
to be useful. The budget here is the smallest scanned one at which the
default keeps at least 30\% of its uncompressed recall. With one task
there is nothing to cluster, so the intervals are prompt-level bootstraps.

\Cref{tab:ledgerdfq} gives the depth curves. The layer picked on RULER
transfers: it beats the default significantly on both per-layer models,
and the curve peaks in the same region as on RULER while the other
family's depth loses, so the good depth belongs to the model family
rather than the task. Picking the layer on the task itself gains
little: the on-task pick, the layer where attention mass on the answer
peaks, ties the RULER pick on Qwen2.5-7B and sits one layer beside it
on Llama-3.1-8B (L13 against L14; chosen on 25 prompts and tested on
the other 175, where it beats the default by $+12.3$pp). On
Gemma-4-E4B-it the mid-depth writer is already the best scorer and
every deeper reader loses, consistent with \cref{sec:shared}, where the
penalty follows the writer's depth; a shared cache has no scorer
shallower than its writer, so only the deep half of the curve is
covered. E2B-it has no entry: no budget up to 2\% damages its writer on
this task.

\begin{table}[!htbp]
\caption{\textbf{The RULER pick transfers to financial-filing
extraction, and a mid-depth writer is already the best scorer of its
shared cache.} DocFinQA test questions on 31.5K-token filing slices;
the metric is recall of the answer's figures; $n{=}200$ per block.
\emph{Top and middle:} Llama-3.1-8B at a 1\% budget and Qwen2.5-7B at
5\%. (pick): the layer the survival model chose on RULER at 32K,
deployed unchanged; (on-task pick): the layer where attention mass on
the answer peaks on these prompts; $^{\dagger}$: a second run on the
same prompts, compared with its own default. \emph{Bottom:}
Gemma-4-E4B-it's shared layer-23 cache at a 1\% budget, scored by the
writer, by one of its readers, or by their consensus; differences from
the writer. Prompt-level 95\%
bootstrap CIs; bold: CI excludes zero.}
\label{tab:ledgerdfq}
\centering\footnotesize\setlength{\tabcolsep}{4pt}
\input{tables/ledger_docfinqa}
\end{table}
\subsection{Statistical checks}\label{app:ledger-signflip}
Clustered intervals use 50k bootstrap draws.
The six primary comparisons are the four combination-over-SnapKV
gains of \cref{tab:faithful} (one per model in column order, the 1.5B at 15\%), the Qwen2.5-0.5B pick (L14) over its
default, and the Gemma-4-E2B best reader over the writer. Each is
tested with an exact sign-flip test over all $2^K$ task-sign patterns
of its $K{=}8$ task means, studentized: $p{=}\sfA$, $\sfB$, $\sfC$,
$\sfD$, $\sfE$, and $\sfF$ in that order (the smallest $p$ the test can give is
$2/2^K{=}\sfFloor$). Because the six share the same eight tasks, a
joint sign-flip on the sum of the six statistics gives $p{=}\sfJoint$,
again the smallest possible. In \cref{tab:shared} the terminal reader's
margin over the writer gives $p{=}\sfLterm$. On the same test, the seven
significant picks of \cref{tab:ledgerpick} give $p{=}0.016$ to $0.062$,
six of them below 0.05.

\section{Eviction, scoring window, and cost}\label{app:physical}
The main-text experiments mask evicted rows rather than deleting them,
which lets the kept set vary per layer and per head. Masking is exact:
softmax with $-\infty$ entries equals softmax over the kept subset. We
checked it by rerunning the single set scored from L19 on Qwen2.5-1.5B
(15\% budget, the eight-task pool, $n{=}200$) with the evicted rows
physically removed: accuracy is identical on all 200 prompts, and 171
generations are token-identical. The other 29 differ from some token
on, as fp16 arithmetic over different tensor shapes allows when two
candidates are nearly tied, and none changes the answer.

\paragraph{Scoring convention.} Every scorer uses attention from a
prefill of the prompt without its final token. The cache is then
compressed, and that final token is decoded under the mask like every
generated token. The kvpress window of \cref{sec:faithful} is therefore the 64 prompt
tokens preceding the final one. GemFilter's statistic is defined on the
final token itself, so \cref{sec:faithful} takes its scores from a
full-prompt prefill.

\paragraph{Runtime and memory.} The \cref{sec:practical} cost numbers come
from Llama-3.1-8B in fp16 on one H200, five prompts per context length,
the first as warm-up and medians over the other four. Scoring as
deployed captures one layer's queries and keys during the ordinary
prefill, which matches a plain prefill within timing noise at both
lengths. The saving arrives at decode: at 32K and a 1\% budget the
decode-resident cache falls from 3.96GB to 45MB and decoding runs
$1.40\times$ faster (81 to 113 tokens/s); at 4K and 5\%, from 0.45GB to
27MB for $1.08\times$. Peak prefill memory is unchanged, because
eviction happens after prefill. On Qwen2.5-1.5B at 4K, scoring from one
layer costs 0.16ms and the top-$k$ selection 0.06ms, together under 1\%
of a prefill that takes 31 to 54ms across prompts; scoring all 28
layers, as the own-layer default does, costs 4.3ms.

\end{document}

%% file: math_commands.tex
\usepackage{amsmath,amsfonts,bm}

\def\eqref#1{equation~\ref{#1}}

\def\1{\bm{1}}

\DeclareMathAlphabet{\mathsfit}{\encodingdefault}{\sfdefault}{m}{sl}
\SetMathAlphabet{\mathsfit}{bold}{\encodingdefault}{\sfdefault}{bx}{n}

%% file: tables/numbers.tex
\newcommand{\sfA}{0.008}
\newcommand{\sfB}{0.016}
\newcommand{\sfC}{0.031}
\newcommand{\sfD}{0.055}
\newcommand{\sfE}{0.031}
\newcommand{\sfF}{0.008}
\newcommand{\sfFloor}{0.008}

\newcommand{\sfJoint}{0.008}
\newcommand{\sfLterm}{0.13}
\newcommand{\eFourGain}{+6.3}
\newcommand{\eFourGainCI}{[-3.2,+17.1]}

\newcommand{\xDesign}{+15.9}
\newcommand{\xDesignCI}{[+4.5,+28.2]}
\newcommand{\gPanelGain}{+36.7}

\newcommand{\spLo}{0.90}
\newcommand{\spHi}{1.00}
\newcommand{\maeLo}{2.7}
\newcommand{\maeHi}{7.4}
\newcommand{\gModR}{0.998}
\newcommand{\gModMax}{5.1}

\newcommand{\lbEtwo}{+8.4}
\newcommand{\lbEtwoCI}{[+1.6,+17.1]}
\newcommand{\lbEfour}{+4.4}

\newcommand{\gMassRatio}{2.4}
\newcommand{\gGapShrink}{6.5}

\newcommand{\gBestHeadLo}{1.5}
\newcommand{\gBestHeadHi}{3.4}
\newcommand{\gfCeil}{90.8}
\newcommand{\gfOurs}{89.3}
\newcommand{\gfGem}{36.5}
\newcommand{\gfGemPeak}{91.4}
\newcommand{\svHalfOwn}{+31.4}

\newcommand{\svFourteenOwn}{+17.7}

\newcommand{\svOneFiveOwn}{+26.6}
\newcommand{\svSmolOwn}{+11.8}
\newcommand{\lcDfqEight}{+13.8}

\newcommand{\lcDfqSeven}{+15.9}

\newcommand{\lcRulerSeven}{+32.5}

\newcommand{\rbHalfSnap}{+40.8}

\newcommand{\rbHalfOwn}{+33.7}

\newcommand{\rbSmolSnap}{+32.6}

\newcommand{\rbSmolOwn}{+15.9}

\newcommand{\consOneAll}{+36.9}

\newcommand{\consOneMid}{+29.9}

\newcommand{\consSmolAll}{+20.8}

\newcommand{\consSmolMid}{+20.0}

%% file: tables/tab1_faithful.tex
\begin{tabular}{lccccc}
\toprule
Scoring method & 1.5B @15\% & 1.5B @10\% & SmolLM2 @15\% & 7B @10\% & 14B @5\% \\
\midrule
No eviction (ceiling) & 91.1 & 90.8 & 76.4 & 93.0 & 91.7 \\
\midrule
H2O (observed attention) & 18.4 & 16.6 & 8.8 & 21.6 & 21.7 \\
SnapKV (per-head, own layer) & 53.0 & 39.2 & 28.4 & 57.4 & 57.6 \\
AdaKV (SnapKV scores) & 57.8 & 42.8 & 30.9 & 53.9 & 53.6 \\
PyramidKV (pyramid budgets) & 33.1 & 22.6 & 20.2 & 28.0 & 34.8 \\
\midrule
Own layer, single set & 63.1 & 49.5 & 48.5 & 77.1 & 56.0 \\
Mid-layer, per-head & 84.7 & 75.1 & 47.0 & 87.2 & 71.6 \\
Mid-layer, single set (ours) & \textbf{87.6} & \textbf{82.0} & \textbf{63.4} & \textbf{91.9} & \textbf{75.9} \\
\midrule
$\Delta$ (ours $-$ SnapKV) & $\mathbf{+34.6}$ & $\mathbf{+42.8}$ & $\mathbf{+34.9}$ & $\mathbf{+34.6}$ & $\mathbf{+18.3}$ \\
\quad 95\% CI & \scriptsize $\mathbf{[+15.6,+54.0]}$ & \scriptsize $\mathbf{[+20.0,+65.8]}$ & \scriptsize $\mathbf{[+14.6,+55.7]}$ & \scriptsize $\mathbf{[+12.6,+57.5]}$ & \scriptsize $\mathbf{[+4.0,+32.8]}$ \\
\bottomrule
\end{tabular}

%% file: tables/tab3_scale.tex
\begin{tabular}{llrlll}
\toprule
Model & Cache & Budget & Best scorer (depth) & Best $-$ own & Worst $-$ own \\
\midrule
Qwen2.5-14B & per-layer & 5\% & L30 (0.64) & \textbf{+17.7 [+5.6,+30.9]} & \textbf{-50.7 [-71.9,-31.5]} \\
Qwen2.5-32B & per-layer & 5\% & L41 (0.65) & +3.9 [-4.2,+11.4] & \textbf{-44.6 [-60.2,-29.4]} \\
Qwen3-0.6B & per-layer & 15\% & L16 (0.59) & \textbf{+20.4 [+5.1,+37.4]} & \textbf{-45.4 [-65.4,-26.9]} \\
Qwen3-1.7B & per-layer & 10\% & L17 (0.63) & +8.3 [-4.6,+21.6] & \textbf{-55.4 [-72.8,-38.5]} \\
Qwen3-14B & per-layer & 5\% & L28 (0.72) & -2.1 [-6.9,+2.4] & \textbf{-42.0 [-63.6,-22.4]} \\
Qwen3.5-9B & hybrid 8/32 & 5\% & L15 (0.48) & +9.8 [-6.2,+26.2] & \textbf{-29.8 [-43.1,-18.7]} \\
Qwen2.5-72B & per-layer & 5\% & L59 (0.75) & +3.1 [-1.6,+8.2] & \textbf{-52.8 [-73.9,-32.7]} \\
Mistral-7B-v0.3 & per-layer & 5\% & L15 (0.48) & \textbf{+9.6 [+3.0,+17.9]} & \textbf{-36.5 [-59.0,-16.6]} \\
Mistral-Small-24B & per-layer & 5\% & L20 (0.51) & +7.2 [-1.5,+18.6] & \textbf{-49.0 [-73.9,-26.2]} \\
Llama-3.2-1B & per-layer & 10\% & L8 (0.53) & \textbf{+24.4 [+5.7,+45.6]} & \textbf{-46.0 [-65.1,-26.9]} \\
Llama-3.1-70B & per-layer & 5\% & L37 (0.47) & +2.4 [-8.5,+16.4] & \textbf{-42.9 [-59.7,-26.1]} \\
Llama-4-Scout & MoE, NoPE 1/4 & 3\% & L32 (0.68) & +7.0 [-2.7,+18.1] & \textbf{-50.0 [-74.3,-26.0]} \\
Llama-3.1-8B (32K) & per-layer & 0.5\% & L14 (0.45) & +7.2 [-6.8,+23.2] & \textbf{-47.8 [-71.5,-25.7]} \\
Qwen2.5-7B (32K) & per-layer & 1\% & L19 (0.70) & +16.3 [-9.8,+42.8] & \textbf{-37.6 [-61.3,-14.9]} \\
\bottomrule
\end{tabular}

%% file: tables/tab2_shared.tex
\begin{tabular}{lcc}
\toprule
Scorer of the shared cache & Accuracy (\%) & $\Delta$ vs.\ writer \\
\midrule
No eviction (ceiling) & 90.3 & -- \\
\midrule
Writer, L14 (default) & 23.1 & -- \\
Reader L19 & 51.2 & $\mathbf{+28.2}$ \scriptsize$\mathbf{[+13.7,+48.0]}$ \\
Reader L24 & 48.3 & $\mathbf{+25.2}$ \scriptsize$\mathbf{[+9.5,+43.6]}$ \\
Reader L29 & 42.5 & $\mathbf{+19.4}$ \scriptsize$\mathbf{[+4.9,+38.7]}$ \\
Reader L34 (terminal) & 37.0 & $\mathbf{+13.9}$ \scriptsize$\mathbf{[+0.8,+33.9]}$ \\
Rank consensus (5 readers) & 45.3 & $\mathbf{+22.2}$ \scriptsize$\mathbf{[+7.6,+39.9]}$ \\
\bottomrule
\end{tabular}

%% file: tables/ledger_binding.tex
\begin{tabular}{lrrrrrrrrrrc}
\toprule
 & \multicolumn{10}{c}{Default damage at each candidate budget (\% of context kept)} & \\
\cmidrule(lr){2-11}
Model & 0.25 & 0.5 & 1 & 2 & 3 & 5 & 10 & 15 & 20 & 25 & Binding \\
\midrule
Qwen2.5-0.5B & -- & -- & -- & -- & -- & 63.2$^{\mathrm{f}}$ & 48.0$^{\mathrm{f}}$ & 44.0$^{\mathrm{f}}$ & 42.6$^{\mathrm{f}}$ & \textbf{38.8} & 25\% \\
Qwen2.5-1.5B & -- & -- & -- & -- & -- & 54.9$^{\mathrm{f}}$ & 34.1$^{\mathrm{f}}$ & \textbf{28.4} & 24.4 & -- & 15\% \\
SmolLM2-1.7B & -- & -- & -- & -- & -- & 46.8$^{\mathrm{f}}$ & 38.2$^{\mathrm{f}}$ & 33.2$^{\mathrm{f}}$ & \textbf{27.4} & -- & 20\% \\
Qwen2.5-7B & -- & -- & -- & -- & -- & 36.6$^{\mathrm{f}}$ & \textbf{13.8} & -- & -- & -- & 10\% \\
Llama-3.2-3B & -- & -- & -- & -- & -- & \textbf{19.0} & 5.2 & -- & -- & -- & 5\% \\
Llama-3.1-8B & -- & -- & -- & -- & -- & \textbf{15.9} & 3.4 & -- & -- & -- & 5\% \\
Gemma-4-E2B (writer) & -- & -- & -- & -- & -- & 84.7$^{\mathrm{f}}$ & 67.4$^{\mathrm{f}}$ & 52.4$^{\mathrm{f}}$ & -- & \textbf{35.2} & 25\% \\
\midrule
Qwen2.5-14B & -- & -- & -- & -- & -- & \textbf{28.0} & 7.2 & 1.2 & -- & -- & 5\% \\
Qwen2.5-32B & -- & -- & -- & -- & -- & \textbf{27.1} & 2.3 & 0.4 & -- & -- & 5\% \\
Qwen3-0.6B & -- & -- & -- & -- & -- & 56.2$^{\mathrm{f}}$ & 39.8$^{\mathrm{f}}$ & \textbf{27.8} & -- & -- & 15\% \\
Qwen3-1.7B & -- & -- & -- & -- & -- & 45.8$^{\mathrm{f}}$ & \textbf{22.4} & 10.3 & -- & -- & 10\% \\
Qwen3-14B & -- & -- & -- & -- & -- & \textbf{39.6} & 11.2 & 6.6 & -- & -- & 5\% \\
Qwen3.5-9B & -- & -- & -- & -- & -- & \textbf{41.2} & 12.0 & 3.9 & -- & -- & 5\% \\
Qwen2.5-72B & -- & -- & -- & -- & -- & \textbf{22.0} & 3.6 & 2.7 & -- & -- & 5\% \\
Mistral-7B-v0.3 & -- & -- & -- & -- & -- & \textbf{31.9} & 6.6 & 0.9 & -- & -- & 5\% \\
Mistral-Small-24B & -- & -- & -- & -- & -- & \textbf{26.0} & 5.5 & 1.5 & -- & -- & 5\% \\
Llama-3.2-1B & -- & -- & -- & -- & -- & 43.3$^{\mathrm{f}}$ & \textbf{24.4} & 16.0 & -- & -- & 10\% \\
Llama-3.1-70B & -- & -- & -- & -- & -- & \textbf{6.8} & -1.0 & -2.5 & -- & -- & 5\% \\
Llama-4-Scout & -- & -- & -- & 49.7$^{\mathrm{f}}$ & \textbf{29.6} & 4.3 & 1.9 & 0.4 & -- & -- & 3\% \\
\midrule
Llama-3.1-8B (32K) & 45.2$^{\mathrm{f}}$ & \textbf{22.6} & 3.3 & 2.4 & -- & 3.1 & 0.6 & -- & -- & -- & 0.5\% \\
Qwen2.5-7B (32K) & 83.3$^{\mathrm{f}}$ & 62.0$^{\mathrm{f}}$ & \textbf{34.2} & 16.8 & -- & 2.8 & -1.1 & -- & -- & -- & 1\% \\
\bottomrule
\end{tabular}

%% file: tables/ledger_picks.tex
\begin{tabular}{lrllll}
\toprule
Model & Budget & Pick & Best other & Pick $-$ own & Pick $-$ best other \\
\midrule
Qwen2.5-0.5B & 15\% & L14 & L15 & \textbf{+31.4 [+9.9,+53.8]} & +10.5 [-0.5,+22.9] \\
Qwen2.5-1.5B & 15\% & L20 & L19 & \textbf{+26.6 [+8.4,+46.5]} & \textbf{+7.4 [+1.8,+14.1]} \\
SmolLM2-1.7B & 15\% & L15 & L16 & \textbf{+11.8 [+3.3,+21.5]} & +0.1 [-6.0,+6.4] \\
Llama-3.2-3B & 5\% & L14 & L16 & +7.2 [-5.8,+22.4] & -1.7 [-13.8,+8.0] \\
Qwen2.5-7B & 10\% & L18 & L19 & +5.5 [-12.5,+26.9] & +1.5 [-3.2,+6.1] \\
Llama-3.1-8B & 5\% & L15 & L13 & +6.0 [-2.9,+16.1] & -1.4 [-5.6,+2.2] \\
Qwen2.5-14B & 5\% & L30 & L29 & \textbf{+17.7 [+5.6,+30.9]} & +5.7 [-2.1,+15.7] \\
Qwen2.5-32B & 5\% & L41 & L50 & +3.9 [-4.2,+11.4] & +1.3 [-7.1,+10.2] \\
Qwen2.5-72B & 5\% & L59 & L55 & +3.1 [-1.6,+8.2] & +3.1 [-2.4,+8.8] \\
Qwen3-0.6B & 15\% & L16 & L15 & \textbf{+20.4 [+5.1,+37.4]} & +5.0 [-0.8,+11.5] \\
Qwen3-1.7B & 10\% & L17 & L15 & +8.3 [-4.6,+21.6] & +4.3 [-3.7,+12.7] \\
Qwen3-14B & 5\% & L28 & L21 & -2.1 [-6.9,+2.4] & +0.7 [-10.8,+12.2] \\
Qwen3.5-9B & 5\% & L15 & L19 & +9.8 [-6.2,+26.2] & +8.3 [-3.9,+19.9] \\
Mistral-7B-v0.3 & 5\% & L15 & L18 & \textbf{+9.6 [+3.0,+17.9]} & \textbf{+9.0 [+2.9,+16.8]} \\
Mistral-Small-24B & 5\% & L28 & L20 & -0.8 [-4.0,+1.8] & -8.0 [-18.9,+0.6] \\
Llama-3.2-1B & 10\% & L8 & L10 & \textbf{+24.4 [+5.7,+45.6]} & +3.0 [-1.2,+7.8] \\
Llama-4-Scout & 3\% & L32 & L26 & +7.0 [-2.7,+18.1] & +3.8 [-0.1,+8.4] \\
Llama-3.1-70B & 5\% & L35 & L37 & +1.5 [-9.7,+15.8] & -0.9 [-3.8,+1.4] \\
\midrule
Qwen2.5-7B (32K) & 1\% & L19 & L22 & +16.3 [-9.8,+42.8] & +5.5 [-14.1,+26.7] \\
Llama-3.1-8B (32K) & 0.5\% & L14 & L20 & +7.2 [-6.8,+23.2] & +1.5 [-4.4,+7.6] \\
\bottomrule
\end{tabular}

%% file: tables/ledger_taskpool.tex
\begin{tabular}{lrlrlrl}
\toprule
 & \multicolumn{2}{c}{Qwen2.5-1.5B @10\%} & \multicolumn{2}{c}{Qwen2.5-1.5B @15\%} & \multicolumn{2}{c}{SmolLM2-1.7B @15\%} \\
\cmidrule(lr){2-3}\cmidrule(lr){4-5}\cmidrule(lr){6-7}
Excluded task & none & ours/SnapKV & none & ours/SnapKV & none & ours/SnapKV \\
\midrule
cwe & 58 & 5/6 & 58 & 12/7 & 4 & 4/10 \\
fwe & 80 & 73/50 & 80 & 67/57 & 70 & 47/40 \\
niah\_multikey\_2 & 100 & 60/20 & 100 & 90/20 & 90 & 10/10 \\
niah\_multikey\_3 & 90 & 10/0 & 90 & 20/10 & 60 & 10/0 \\
niah\_single\_3 & 100 & 0/0 & 100 & 0/0 & 100 & 0/0 \\
\midrule
ours $-$ SnapKV: five excluded tasks & \multicolumn{2}{c}{\textbf{+14.5 [+0.8,+32.0]}} & \multicolumn{2}{c}{\textbf{+19.0 [+2.0,+46.4]}} & \multicolumn{2}{c}{+4.2 [-5.8,+15.0]} \\
ours $-$ SnapKV: eight-task pool & \multicolumn{2}{c}{\textbf{+42.8 [+20.0,+65.8]}} & \multicolumn{2}{c}{\textbf{+34.6 [+15.6,+54.0]}} & \multicolumn{2}{c}{\textbf{+34.9 [+14.6,+55.7]}} \\
ours $-$ SnapKV: all thirteen tasks & \multicolumn{2}{c}{\textbf{+37.1 [+17.7,+56.1]}} & \multicolumn{2}{c}{\textbf{+31.5 [+15.0,+47.7]}} & \multicolumn{2}{c}{\textbf{+29.8 [+11.6,+47.4]}} \\
\bottomrule
\end{tabular}

%% file: tables/ledger_tab1_more.tex
\begin{tabular}{lccccc}
\toprule
Scoring method & 1.5B @5\% & 1.5B @20\% & 0.5B @15\% & 3B @5\% & 8B @5\% \\
\midrule
No eviction (ceiling) & 90.8 & 90.8 & 84.6 & 91.3 & 93.9 \\
\midrule
H2O (observed attention) & 13.7 & 19.9 & 15.5 & 17.4 & 21.1 \\
SnapKV (per-head, own layer) & 17.9 & 60.5 & 35.1 & 74.7 & 78.6 \\
AdaKV (SnapKV scores) & 20.1 & 66.8 & 31.4 & 74.0 & 75.1 \\
PyramidKV (pyramid budgets) & 10.4 & 44.8 & 22.7 & 62.2 & 72.3 \\
\midrule
Own layer, single set & 23.1 & 73.4 & 31.9 & 74.6 & 78.2 \\
Mid-layer, per-head & 48.5 & 86.4 & \textbf{74.0} & 72.0 & \textbf{81.4} \\
Mid-layer, single set (ours) & \textbf{55.7} & \textbf{88.3} & 72.1 & \textbf{76.8} & 81.0 \\
\midrule
$\Delta$ (ours $-$ SnapKV) & $\mathbf{+37.8}$ & $\mathbf{+27.7}$ & $\mathbf{+37.0}$ & $+2.1$ & $+2.4$ \\
\quad 95\% CI & \scriptsize $\mathbf{[+16.1,+61.0]}$ & \scriptsize $\mathbf{[+9.5,+46.2]}$ & \scriptsize $\mathbf{[+15.3,+58.9]}$ & \scriptsize $[-9.0,+13.6]$ & \scriptsize $[-4.8,+9.3]$ \\
\bottomrule
\end{tabular}

%% file: tables/ledger_tab1_loose.tex
\begin{tabular}{lccc}
\toprule
Scoring method & 1.5B @30\% & 1.5B @40\% & 1.5B @50\% \\
\midrule
No eviction (ceiling) & 90.8 & 90.8 & 90.8 \\
\midrule
H2O (observed attention) & 26.6 & 33.2 & 40.7 \\
SnapKV (per-head, own layer) & 70.0 & 78.1 & 82.4 \\
AdaKV (SnapKV scores) & 76.1 & 81.8 & 84.9 \\
PyramidKV (pyramid budgets) & 63.7 & 69.6 & 79.2 \\
\midrule
Own layer, single set & 81.3 & 84.9 & 85.0 \\
Mid-layer, per-head & 88.3 & 89.8 & 89.9 \\
Mid-layer, single set (ours) & \textbf{89.5} & \textbf{90.5} & \textbf{90.0} \\
\midrule
$\Delta$ (ours $-$ SnapKV) & $\mathbf{+19.4}$ & $\mathbf{+12.5}$ & $\mathbf{+7.6}$ \\
\quad 95\% CI & \scriptsize $\mathbf{[+6.3,+34.4]}$ & \scriptsize $\mathbf{[+4.0,+22.6]}$ & \scriptsize $\mathbf{[+0.8,+16.1]}$ \\
\bottomrule
\end{tabular}

%% file: tables/ledger_tab1_32k.tex
\begin{tabular}{lcccc}
\toprule
 & \multicolumn{2}{c}{DocFinQA, 31.5K} & \multicolumn{2}{c}{RULER, 32K} \\
\cmidrule(lr){2-3}\cmidrule(lr){4-5}
Scoring method & 8B @1\% & 7B @5\% & 8B @0.5\% & 7B @1\% \\
\midrule
No eviction (ceiling) & 45.9 & 46.2 & 91.7 & 88.3 \\
\midrule
SnapKV (per-head, own layer) & 15.0 & 21.6 & 68.3 & 40.1 \\
AdaKV (SnapKV scores) & 14.6 & 19.9 & 64.6 & 37.1 \\
PyramidKV (pyramid budgets) & 15.1 & 14.7 & 62.0 & 21.4 \\
\midrule
Own layer, single set & 17.6 & 24.6 & 68.8 & 51.2 \\
Mid-layer, per-head & 24.0 & 22.9 & \textbf{74.2} & 62.0 \\
Mid-layer, single set (ours) & \textbf{28.8} & \textbf{37.5} & 72.9 & \textbf{72.6} \\
\midrule
$\Delta$ (ours $-$ SnapKV) & $\mathbf{+13.8}$ & $\mathbf{+15.9}$ & $+4.6$ & $\mathbf{+32.5}$ \\
\quad 95\% CI & \scriptsize $\mathbf{[+9.2,+18.6]}$ & \scriptsize $\mathbf{[+11.0,+21.0]}$ & \scriptsize $[-3.7,+12.3]$ & \scriptsize $\mathbf{[+8.1,+58.5]}$ \\
\bottomrule
\end{tabular}

%% file: tables/ledger_tab1_parts.tex
\begin{tabular}{lccccc}
\toprule
$\Delta$ vs.\ SnapKV & 1.5B @15\% & 1.5B @10\% & SmolLM2 @15\% & 7B @10\% & 14B @5\% \\
\midrule
single set (own layer) & $\mathbf{+10.1}$ & $\mathbf{+10.3}$ & $\mathbf{+20.1}$ & $\mathbf{+19.7}$ & $-1.4$ \\
\quad 95\% CI & \scriptsize $\mathbf{[+2.7,+18.9]}$ & \scriptsize $\mathbf{[+1.4,+23.0]}$ & \scriptsize $\mathbf{[+1.1,+42.3]}$ & \scriptsize $\mathbf{[+8.9,+31.9]}$ & \scriptsize $[-9.3,+7.0]$ \\
mid layer (per-head) & $\mathbf{+31.7}$ & $\mathbf{+35.9}$ & $\mathbf{+18.6}$ & $\mathbf{+29.8}$ & $\mathbf{+14.0}$ \\
\quad 95\% CI & \scriptsize $\mathbf{[+14.6,+49.5]}$ & \scriptsize $\mathbf{[+17.6,+55.9]}$ & \scriptsize $\mathbf{[+0.9,+40.1]}$ & \scriptsize $\mathbf{[+10.0,+51.1]}$ & \scriptsize $\mathbf{[+0.4,+26.7]}$ \\
AdaKV & $\mathbf{+4.8}$ & $+3.6$ & $+2.5$ & $-3.5$ & $-3.9$ \\
\quad 95\% CI & \scriptsize $\mathbf{[+0.5,+10.1]}$ & \scriptsize $[-2.0,+10.5]$ & \scriptsize $[-7.1,+13.6]$ & \scriptsize $[-12.0,+3.7]$ & \scriptsize $[-13.1,+3.3]$ \\
\bottomrule
\end{tabular}

%% file: tables/ledger_pertask.tex
\begin{tabular}{lrrrrrrr}
\toprule
 & \multicolumn{5}{c}{4K (\cref{tab:faithful})} & \multicolumn{2}{c}{32K (\cref{tab:ledgerlong})} \\
\cmidrule(lr){2-6}\cmidrule(lr){7-8}
Task & 1.5B @15\% & 1.5B @10\% & SmolLM2 @15\% & 7B @10\% & 14B @5\% & 8B @0.5\% & 7B @1\% \\
\midrule
niah\_single\_1 & +24 & +52 & +80 & +12 & +16 & +0 & +12 \\
niah\_single\_2 & +52 & +84 & +68 & +52 & +44 & +0 & +88 \\
niah\_multikey\_1 & +72 & +80 & +40 & +72 & +44 & +0 & +56 \\
niah\_multiquery & +65 & +79 & +37 & +49 & +27 & +12 & +83 \\
niah\_multivalue & +51 & +23 & +6 & +84 & +3 & +20 & +20 \\
\midrule
vt & +2 & +10 & +47 & +13 & +27 & +17 & +17 \\
\midrule
qa\_1 & +3 & +3 & +1 & -1 & -7 & -12 & -12 \\
qa\_2 & +8 & +12 & +0 & -4 & -8 & +0 & -4 \\
\bottomrule
\end{tabular}

%% file: tables/ledger_longbench.tex
\begin{tabular}{lrrr}
\toprule
Scorer & F1 & $\Delta$ vs.\ default & 95\% CI \\
\midrule
\multicolumn{4}{l}{\emph{Qwen2.5-1.5B, 5\% (binding); 8 of 9 datasets informative}} \\
\midrule
none (no eviction) & 36.6 & -- & -- \\
own (per-layer default) & 30.2 & -- & -- \\
L2 & 24.4 & \textbf{-5.8} & \textbf{[-12.0,-0.3]} \\
L19 & 30.0 & -0.2 & [-3.7,+3.1] \\
L20 & 29.4 & -0.9 & [-4.7,+2.5] \\
L27 & 25.1 & \textbf{-5.2} & \textbf{[-10.6,-0.7]} \\
\midrule
\multicolumn{4}{l}{\emph{Llama-3.1-8B, 2\%; 9 of 9 datasets informative}} \\
\midrule
none (no eviction) & 43.5 & -- & -- \\
own (per-layer default) & 38.7 & -- & -- \\
L2 & 32.6 & \textbf{-6.1} & \textbf{[-11.2,-1.8]} \\
L13 & 36.9 & -1.8 & [-5.8,+1.3] \\
L14 & 37.8 & -0.9 & [-5.2,+3.4] \\
L17 & 37.7 & -1.0 & [-4.6,+2.4] \\
L31 & 33.4 & \textbf{-5.3} & \textbf{[-10.5,-0.9]} \\
\bottomrule
\end{tabular}

%% file: tables/tab2_shared_reps.tex
\begin{tabular}{lc}
\toprule
Replication & reader $-$ writer \\
\midrule
reader L19 $-$ writer @ budget 5\% & $\mathbf{+16.8}$ \scriptsize$\mathbf{[+1.5,+38.0]}$ \\
reader L19 $-$ writer @ budget 15\% & $\mathbf{+36.3}$ \scriptsize$\mathbf{[+16.0,+58.1]}$ \\
reader L19 $-$ writer @ budget 25\% & $\mathbf{+30.2}$ \scriptsize$\mathbf{[+12.0,+48.7]}$ \\
reader L19 $-$ writer @ 8K context & $\mathbf{+36.5}$ \scriptsize$\mathbf{[+13.9,+61.4]}$ \\
reader L19 $-$ writer @ 16K context & $\mathbf{+33.8}$ \scriptsize$\mathbf{[+15.5,+52.0]}$ \\
Gemma-3n: reader L24 $-$ writer L19 & $\mathbf{+12.5}$ \scriptsize$\mathbf{[+3.2,+23.2]}$ \\
\bottomrule
\end{tabular}

%% file: tables/ledger_gemma.tex
\begin{tabular}{lrrrrll}
\toprule
Replication & none & writer & L19 & cons. & L19 $-$ writer & cons.\ $-$ writer \\
\midrule
disjoint prompt draw (5 tasks) & 99.0 & 24.8 & -- & 57.1 & -- & \textbf{+32.3 [+10.0,+56.4]} \\
13 tasks, base & 90.6 & 22.9 & 41.0 & 40.9 & \textbf{+18.1 [+4.6,+35.1]} & \textbf{+18.0 [+4.6,+33.8]} \\
13 tasks, instruct & 84.9 & 39.3 & 63.9 & 51.1 & \textbf{+24.6 [+10.0,+41.2]} & \textbf{+11.8 [+3.5,+21.9]} \\
\bottomrule
\end{tabular}

%% file: tables/ledger_rh.tex
\begin{tabular}{lrlll}
\toprule
Model & Budget & Heads $-$ own & Layer & Heads $-$ layer \\
\midrule
Qwen2.5-1.5B & 15\% & \textbf{+26.0 [+6.4,+47.8]} & L19 (grid) & \textbf{+6.8 [+0.2,+15.0]} \\
 & & & L20 (pick) & -0.6 [-3.7,+2.0] \\
Llama-3.1-8B & 5\% & \textbf{+9.0 [+1.1,+17.8]} & L13 & +1.4 [-1.6,+5.0] \\
Mistral-Small-24B & 5\% & \textbf{+18.7 [+2.0,+38.5]} & L17 & \textbf{+11.5 [+3.5,+20.1]} \\
\bottomrule
\end{tabular}

%% file: tables/ledger_docfinqa.tex
\begin{tabular}{lrrl}
\toprule
Scorer & Depth & Recall (\%) & $\Delta$ vs.\ default \\
\midrule
\multicolumn{4}{l}{\emph{Llama-3.1-8B, per-layer caches (1\%, $n{=}200$)}} \\
\midrule
none (no eviction) & -- & 45.9 & -- \\
own (per-layer default) & -- & 25.2 & -- \\
\midrule
L2 & 0.06 & 1.4 & \textbf{-23.8 [-28.8,-18.9]} \\
L8 & 0.26 & 3.2 & \textbf{-22.0 [-26.9,-17.5]} \\
L12 & 0.39 & 1.9 & \textbf{-23.3 [-28.4,-18.5]} \\
L13 (on-task pick) & 0.42 & 37.0 & \textbf{+11.8 [+7.0,+16.6]} \\
L14 (pick) & 0.45 & 33.4 & \textbf{+8.2 [+4.0,+12.4]} \\
L17 & 0.55 & 29.5 & \textbf{+4.2 [+0.1,+8.7]} \\
L22 & 0.71 & 15.0 & \textbf{-10.2 [-14.9,-5.6]} \\
L26 & 0.84 & 8.9 & \textbf{-16.3 [-21.5,-11.3]} \\
L31 & 1.00 & 2.2 & \textbf{-23.0 [-28.1,-18.0]} \\
\midrule
\multicolumn{4}{l}{\emph{Qwen2.5-7B, per-layer caches (5\%, $n{=}200$)}} \\
\midrule
none (no eviction) & -- & 46.5 & -- \\
own (per-layer default) & -- & 24.4 & -- \\
\midrule
L2 & 0.07 & 2.7 & \textbf{-21.8 [-27.0,-16.7]} \\
L7 & 0.26 & 2.4 & \textbf{-22.0 [-27.3,-17.0]} \\
L11 & 0.41 & 6.7 & \textbf{-17.8 [-23.0,-12.7]} \\
L15 & 0.56 & 12.3 & \textbf{-12.1 [-16.9,-7.5]} \\
L16$^{\dagger}$ (on-task pick) & 0.59 & 40.9 & \textbf{+16.2 [+11.3,+21.2]} \\
L17 & 0.63 & 34.3 & \textbf{+9.9 [+5.3,+14.5]} \\
L18$^{\dagger}$ & 0.67 & 34.5 & \textbf{+9.8 [+4.7,+15.0]} \\
L19 (pick) & 0.70 & 36.6 & \textbf{+12.2 [+8.0,+16.8]} \\
L21 & 0.78 & 32.5 & \textbf{+8.1 [+4.4,+12.1]} \\
L23 & 0.85 & 25.3 & +0.9 [-2.6,+4.4] \\
L27 & 1.00 & 6.2 & \textbf{-18.2 [-23.1,-13.5]} \\
\midrule
\multicolumn{4}{l}{\emph{Gemma-4-E4B-it, shared layer-23 cache (1\%, $n{=}200$)}} \\
\midrule
none (no eviction) & -- & 48.9 & -- \\
writer L23 (default) & 0.56 & 21.8 & -- \\
\midrule
L29 & 0.71 & 8.0 & \textbf{-13.7 [-18.0,-9.6]} \\
L35 & 0.85 & 3.2 & \textbf{-18.6 [-23.8,-13.8]} \\
L41 & 1.00 & 0.2 & \textbf{-21.5 [-26.6,-16.7]} \\
consensus (4 readers) & -- & 6.3 & \textbf{-15.4 [-20.3,-10.8]} \\
\bottomrule
\end{tabular}